\documentclass{article} 
\usepackage{iclr2026_conference,times}

\usepackage{amsmath,amsfonts,bm}

\def\eqref#1{equation~\ref{#1}}

\def\1{\bm{1}}

\DeclareMathAlphabet{\mathsfit}{\encodingdefault}{\sfdefault}{m}{sl}
\SetMathAlphabet{\mathsfit}{bold}{\encodingdefault}{\sfdefault}{bx}{n}

\usepackage{hyperref}
\usepackage{url}

\usepackage{booktabs}       
\usepackage{amsfonts}       
\usepackage{nicefrac}       
\usepackage{microtype}      
\usepackage{xcolor}         
\usepackage{graphicx}
\usepackage{subfigure} 
\usepackage{wrapfig}
\usepackage{amsmath}
\usepackage{multirow}
\usepackage[table]{xcolor}

\title{L2P: Unlocking Latent Potential for Pixel Generation}

\author{Zhennan Chen$^{1,2}$\thanks{Work done during the internship at Tencent. \\$^\dagger$Project Leader. $^\ddagger$Corresponding Author.}  ~ Junwei Zhu$^2$$^\dagger$ ~ Xu Chen$^{2}$ ~ Jiangning Zhang$^2$ ~ Jiawei Chen$^{1}$ \\ \textbf{Zhuoqi Zeng$^3$} ~ \textbf{Wei Zhang$^4$} \textbf{Chengjie Wang$^2$} ~ \textbf{Jian Yang$^1$} ~ \textbf{Ying Tai$^1$$^\ddagger$} \\
	$^1$Nanjing University  ~~  $^2$Tencent Youtu Lab~~ $^3$Hainan-biuh~~ $^4$Weess Gmbh~~\\
    {\small \textcolor{magenta}{\url{https://nju-pcalab.github.io/projects/L2P/}}}
}

\iclrfinalcopy 
\begin{document}

\maketitle

\begin{abstract}
Pixel diffusion models have recently regained attention for visual generation. However, training advanced pixel-space models from scratch demands prohibitive computational and data resources. To address this, we propose the \textbf{Latent-to-Pixel (L2P)} transfer paradigm, an efficient framework that directly harnesses the rich knowledge of pre-trained LDMs to build powerful pixel-space models. Specifically, L2P discards the VAE in favor of large-patch tokenization and freezes the source LDM's intermediate layers, exclusively training shallow layers to learn the latent-to-pixel transformation. By utilizing LDM-generated synthetic images as the sole training corpus, L2P fits an already smooth data manifold, enabling rapid convergence with zero real-data collection. This strategy allows L2P to seamlessly migrate massive latent priors to the pixel space using only 8 GPUs. Furthermore, eliminating the VAE memory bottleneck unlocks native 4K ultra-high resolution generation. Extensive experiments across mainstream LDM architectures show that L2P incurs negligible training overhead, yet performs on par with the source LDM on DPG-Bench and reaches 93\% performance on GenEval.

\end{abstract}

\begin{wrapfigure}{r}{7cm}
    \centering
    \vspace{-6mm}
    \includegraphics[width=\linewidth]{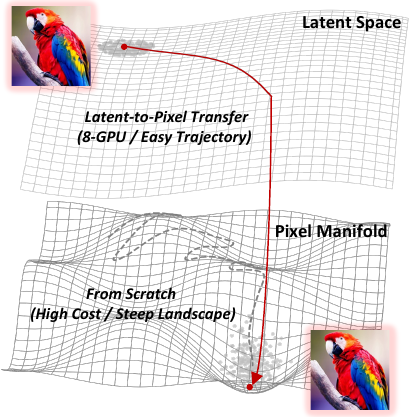}
    \vspace{-5.5mm}
    \caption{By leveraging the smooth manifold of pre-trained LDMs, L2P bypasses costly from-scratch training, achieving high-quality generation with just 8 GPUs.}
    \vspace{-6mm}
    \label{Fig:teaser}
\end{wrapfigure}

\section{Introduction}
\label{sec:intro}

Latent Diffusion Models (LDMs)~\cite{sohl2015deep, ho2020denoising, song2020score, peebles2023scalable,ramesh2022hierarchical,saharia2022photorealistic,yu2022scaling, xie2024sana,song2020denoising,ho2022classifier,karras2024guiding} have recently dominated the field of text-to-image (T2I) generation~\cite{cai2025z,wu2025qwen,wang2024instantid, chen2025ragd, zhou2024migc,zhou20243dis,chen2023pixart,du2025textcrafter}, achieving unprecedented success in synthesizing high-quality images. By compressing images into a lower-dimensional latent space via a Variational Autoencoder (VAE)~\cite{kingma2013auto}, LDMs significantly reduce computational overhead. 
Nevertheless, this bipartite paradigm is inherently bounded by VAE-induced limitations. The compression process inevitably discards critical high-frequency details~\cite{cai2026vae,yao2025reconstruction,kilian2024computational,chen2024deep,gupta2024photorealistic}, leading to sub-optimal reconstruction and a non-end-to-end training pipeline that decouples representation learning from the generation process.
Furthermore, the VAE decoding process imposes severe memory constraints, bottlenecking the scaling to ultra-high resolutions (e.g., native 4K). To circumvent these VAE-induced limitations and achieve uncompromised visual fidelity, pixel-space diffusion models have recently re-emerged as a promising alternative~\cite{chen2025dip,li2025jit,ma2025deco,wang2025pixnerd,ma2026pixelgen,yu2025pixeldit}.

Despite their architectural purity and end-to-end appeal, training a state-of-the-art pixel-space T2I model from scratch remains computationally prohibitive, typically demanding hundreds of high-end GPUs and billions of curated image-text pairs. Consequently, nascent pixel-space models~\cite{ma2025deco,wang2025pixnerd,ma2026pixelgen,yu2025pixeldit} frequently exhibit a pronounced gap in semantic comprehension and compositional quality when compared to established LDMs~\cite{cai2025z,wu2025qwen,esser2024scaling,flux}, which have already internalized profound world knowledge distilled from massive-scale datasets. This presents a critical cold-start dilemma: \textit{Can we directly transfer the rich semantic priors embedded in pre-trained LDMs to a pixel-space diffusion model, thereby bypassing the astronomical costs of from-scratch training?}

To this end, we propose the \textbf{Latent-to-Pixel (L2P)} transfer paradigm, a highly efficient framework designed to bridge the representation gap between latent and pixel spaces at low cost, as shown in Figure~\ref{Fig:teaser}. Architecturally, we discard the VAE, employ large-patch tokenization for pixel inputs, and utilize a lightweight U-Net to manage the decoding process. To facilitate robust knowledge transfer, we keep the Diffusion Transformer (DiT) architecture unmodified and align the prediction target with the source LDM. This architectural fidelity ensures seamless weight inheritance, while objective consistency allows the frozen intermediate layers to function within their native optimization manifold, thereby preserving the rich semantic priors and world knowledge. Consequently, we freeze the intermediate layers of the DiT backbone and exclusively train the shallow input and output layers to learn the latent-to-pixel modality transformation. Furthermore, rather than collecting massive real-world datasets, we utilize the source LDM to generate high-quality images as our training corpus. Beyond eliminating data curation costs, this strategy forces the new pixel model to fit the smooth data manifold already constructed by the LDM, thereby drastically accelerating convergence. Moreover, eliminating the VAE bottleneck unlocks native 4K generation. We maintain computational efficiency at this scale simply by enlarging the patch size and increasing the noise shift. The resulting heavier noise fully corrupts the dense local correlations of 4K pixels, averting trivial local reconstruction and enforcing global structural learning. 

Our contributions are summarized as follows:

\noindent$\bullet$ We propose Latent-to-Pixel (L2P), a highly resource-efficient transfer paradigm that harnesses massive pre-trained LDM priors for pixel-space diffusion using merely 8 GPUs, seamlessly transitioning to the pixel space while simultaneously unlocking native 4K ultra-high-resolution generation.

\noindent$\bullet$ We construct a comprehensive, multi-dimensional prompt dataset to generate synthetic training pairs, achieving highly efficient training with zero real-data cost.

\noindent$\bullet$ Extensive validations demonstrate that L2P robustly inherits the generative priors of the source LDM. It maintains near-lossless semantic alignment on standard benchmarks while simultaneously exhibiting exceptional visual fidelity in native 4K ultra-high-resolution generation.
\section{Related Work}
\label{sec: related_work}

\textbf{Text-to-Image Generation.}
Text-to-Image (T2I) generation~\cite{podell2023sdxl,chen2023diffusion,ye2023ip,wang2024instantid,zhao2025ultrahr, chen2025ragd, zhou2024migc,zhou20243dis,zhao2024wavelet,chen2023pixart,gao2025subject, dong2025vita,du2025textcrafter,zhou2026refineanything, zhao2026learning} is currently dominated by LDMs~\cite{rombach2022high}, which bypass the exorbitant computational costs of early pixel-space models~\cite{dhariwal2021diffusion, ho2020denoising} by compressing images into a compact latent space via a Variational Autoencoder (VAE)~\cite{kingma2013auto}. Despite encapsulating profound world knowledge and robust semantic alignment, LDMs are inherently bottlenecked by the VAE decoder. The compression-decompression process inevitably incurs high-frequency information loss~\cite{yao2025reconstruction,kilian2024computational,chen2024deep,gupta2024photorealistic}. Furthermore, the severe quadratic memory footprint of the VAE spatial decoding process imposes rigid hardware constraints, making native ultra-high resolution (e.g., 4K) generation practically intractable for standard LDMs~\cite{zhao2025ultrahr,chen2024pixart,zhang2025diffusion,xie2024sana,du2024max,bu2025hiflow,zhao2026zero,pixverve}.

\textbf{Pixel Diffusion Models.} 
Early pixel diffusion models (e.g., DDPM~\cite{ho2020denoising} and ADM~\cite{dhariwal2021diffusion}) are severely constrained when processing high-resolution images due to their quadratic complexity bottleneck. Approaches like JiT~\cite{li2025jit} and PixelGen~\cite{ma2026pixelgen} introduce novel prediction targets. Most relevantly, PixNerd~\cite{wang2025pixnerd}, DeCo~\cite{ma2025deco}, PixelDiT~\cite{yu2025pixeldit}, and DiP~\cite{chen2025dip} efficiently decouple global structural modeling from local detail refinement via lightweight decoders. Despite their architectural advances, these modern models still mandate computationally prohibitive from-scratch training on massive datasets. In contrast, our work fundamentally circumvents these exorbitant pre-training costs. Through our L2P paradigm, we directly transfer the rich priors of existing LDMs into the pixel space, achieving state-of-the-art pixel-based text-to-image generation with minimal computational overhead.

\section{Method}

\subsection{Preliminary}

Diffusion models learn to synthesize data by reversing a progressive noise-injection process. Given an initial sample $\mathbf{x}_{0} \sim q(\mathbf{x}_{0})$, the discrete forward process yields a noisy state at step $t$:
\begin{equation}
    \mathbf{x}_{t} = \sqrt{\bar{\alpha}_{t}} \mathbf{x}_{0} + \sqrt{1-\bar{\alpha}_{t}} \epsilon, \quad \epsilon \sim \mathcal{N}(0, \mathbf{I}),
\end{equation}
where $\bar{\alpha}_{t}$ is determined by a predefined variance schedule. As $t \rightarrow T$, the marginal distribution $p(\mathbf{x}_{T})$ converges to a standard Gaussian $\mathcal{N}(0, \mathbf{I})$. 
In a continuous-time framework, this corruption process is governed by a stochastic differential equation (SDE) $d\mathbf{x} = f(\mathbf{x}, t)dt + g(t)d\mathbf{w}$, with drift $f(\cdot, t)$ and diffusion coefficient $g(t)$. The generative process corresponds to simulating the reverse-time Probability Flow ODE:
\begin{equation}
    d\mathbf{x} = \left[ f(\mathbf{x}, t) - \frac{1}{2}g(t)^{2} \nabla_{\mathbf{x}} \log p_{t}(\mathbf{x}) \right] dt.
\end{equation}
Consequently, data generation relies on estimating the score function $\nabla_{\mathbf{x}} \log p_{t}(\mathbf{x})$ or the associated vector field. A standard approach (e.g., DDPM) trains a neural network $\epsilon_{\theta}(\mathbf{x}_{t}, t)$ to predict the injected noise:
\begin{equation}
    \mathcal{L}_{\mathrm{DDPM}} = \mathbb{E}_{t, \mathbf{x}_{0}, \epsilon} \left[ \left\| \epsilon - \epsilon_{\theta}(\mathbf{x}_{t}, t) \right\|^{2} \right].
\end{equation}
Alternatively, Flow Matching (FM)~\cite{esser2024scaling} offers a simulation-free paradigm to directly regress the continuous vector field. By defining a conditional probability path $p_{t}(\mathbf{x} \mid \mathbf{x}_{0})$ and its target vector field $u_{t}(\mathbf{x})$, a model $v_{\theta}(\mathbf{x}, t)$ is optimized via:
\begin{equation}
    \mathcal{L}_{\mathrm{FM}} = \mathbb{E}_{t, p_{t}(\mathbf{x} \mid \mathbf{x}_{0})} \left[ \left\| u_{t}(\mathbf{x}) - v_{\theta}(\mathbf{x}, t) \right\|^{2} \right].
\end{equation}

\begin{figure*}[!ht]
    \centering
    \includegraphics[width=\textwidth]{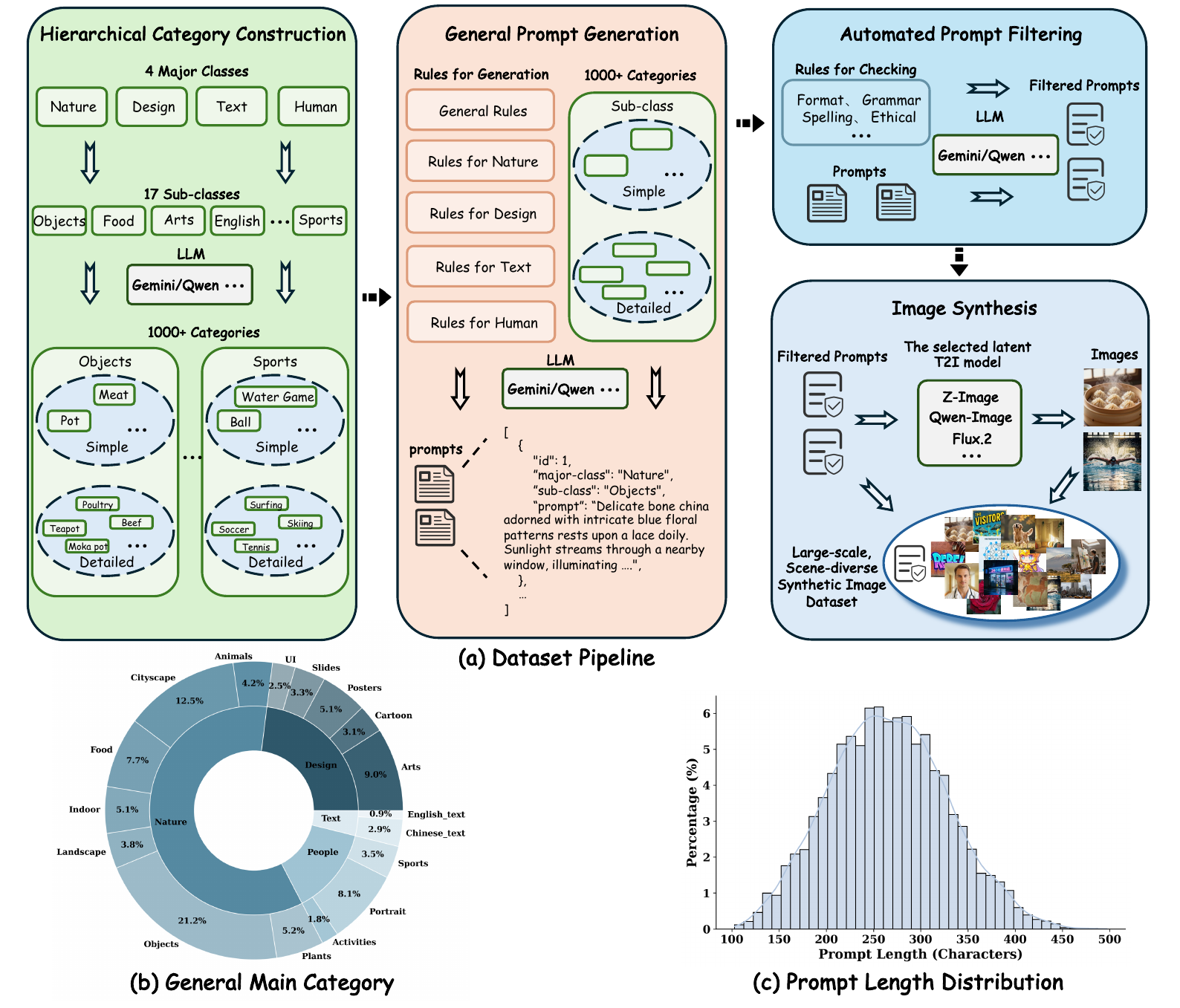}
    \vspace{-4mm}
    \caption{The proposed data construction pipeline. (a) Four-stage construction framework: Hierarchical Category Construction, General Prompt Generation, Automated Prompt Filtering, and Image Synthesis. Further details are provided in the Appendix. (b) Category distribution detailing 4 super-classes and 17 sub-classes. (c) Prompt length distribution, peaking at 200–350 characters to provide rich textual details.}
    \label{Fig:dataset_pipeline}
    \vspace{-4mm}
\end{figure*}

\subsection{Dataset Construction}
\label{sec:data_construction}

To facilitate the L2P transfer without the prohibitive costs of real-world data collection, we designed a comprehensive dataset pipeline, as shown in Figure~\ref{Fig:dataset_pipeline}(a). Through this pipeline, we construct a large-scale, scene-diverse synthetic image dataset. Generating our training corpus directly from the source LDM forces the new pixel-space model to fit the smooth data manifold already constructed by the source model, significantly accelerating convergence and activating its intrinsic prior knowledge. Our data construction process is structured into the following sequential stages:

\textbf{Hierarchical Category Construction.} To ensure comprehensive semantic coverage and diversity, we establish a top-down hierarchical taxonomy. First, drawing upon~\cite{wu2025qwen,team2025longcat}, we define 4 major classes and further divide them into 17 sub-classes, as shown in Figure~\ref{Fig:dataset_pipeline}(b). Subsequently, we leverage an LLM to expand these sub-classes into over 1,000 fine-grained categories.

\textbf{General Prompt Generation.} We design a refined set of generation rules to guide the LLM in synthesizing high-quality prompts. Guided by these customized rules and the 1,000+ categories, the LLM generates highly descriptive prompts formatted as structured JSON data. As shown in Figure~\ref{Fig:dataset_pipeline}(c), the generated prompts are densely concentrated between 200 and 350 characters, providing abundant textual details for complex scene generation.

\textbf{Automated Prompt Filtering.} To prevent the propagation of low-quality or unsafe data, we implement a rigorous prompt check. The rules for check filter the generated text based on strict criteria. This ensures a high-quality corpus of filtered prompts.

\textbf{Image Synthesis.} Finally, for image generation, we feed the filtered prompts into the source latent T2I model to synthesize the final images.

\subsection{L2P Transfer Paradigm}
\begin{figure*}[h]
    \centering
    \includegraphics[width=\textwidth]{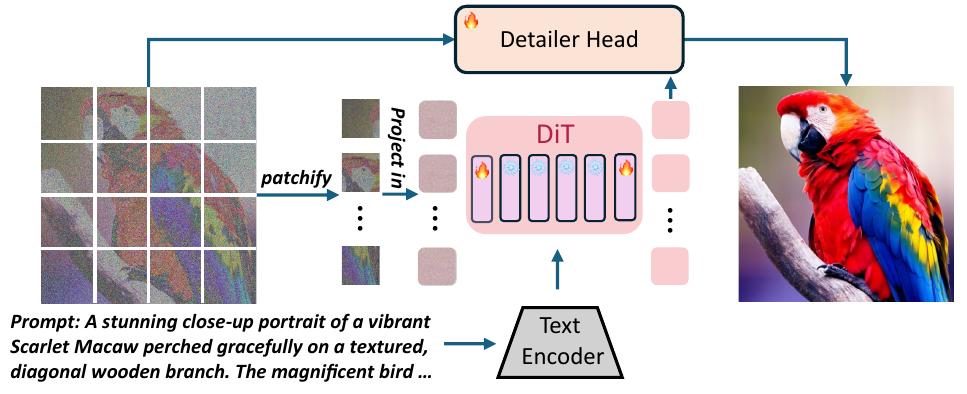}
    \vspace{-4mm}
    \caption{Overview of the L2P framework. L2P operates directly in pixel space via large-patch tokenization without VAE. To efficiently adapt priors, core DiT layers are frozen, while shallow blocks and a Detailer Head are tuned to restore high-frequency spatial details.
    }
    \label{Fig:L2P}
    \vspace{-4mm}
\end{figure*}

To efficiently migrate the rich generative priors embedded in pre-trained LDMs into the pixel space, we introduce the L2P transfer paradigm. The overall architecture is illustrated in Figure~\ref{Fig:L2P}.

\textbf{Architectural Adaptation.} To facilitate the transition from latent to pixel space without disrupting the internal sequence processing of the pre-trained Diffusion Transformer (DiT), we implement three structural modifications:

\textit{1)} We discard the VAE and apply a patchification strategy to the input image. To align the sequence length and maintain the computational efficiency equivalent to the original VAE-compressed latent space, we employ a patch size of 16$\times$16. 

\textit{2)} Pre-trained LDMs map latent representations back to images via a VAE decoder. To bypass the VAE decoder bottleneck and enable high-fidelity pixel-level generation, inspired by DiP~\cite{chen2025dip}, we replace the final projection layer with a lightweight U-Net, termed the Detailer Head. This module decodes DiT representations to reconstruct dense pixel semantics and restore high-frequency details.

\textit{3)} To achieve rapid convergence while preventing catastrophic forgetting of the LDM's semantic priors, we employ a selective freezing strategy. During training, the majority of the intermediate DiT blocks are frozen. We only update the initial input projection layer, the first and last $n$ blocks of the DiT, and the newly added Detailer Head. This drastically reduces the computational overhead compared to training from scratch.

\textbf{Objective Function.} To maximize the preservation of pre-trained generative priors, we strictly adhere to the original diffusion training objective of the source LDM. The L2P optimization objective is formulated as:
\begin{equation}
    \mathcal{L}_{\mathrm{L2P}} = \mathbb{E}_{\mathbf{x}_0, \epsilon, t} \left[ \left\| (\epsilon - \mathbf{x}_0) - v_{\theta}(\mathbf{x}_t, t) \right\|^{2} \right]
\end{equation}
By maintaining optimization consistency with the source model, L2P inherently mitigates the catastrophic forgetting of pre-trained knowledge. Furthermore, this architecture-agnostic formulation ensures seamless deployment across diverse LDM frameworks.

\begin{wrapfigure}{r}{6cm}
    \centering
    \vspace{-10mm}
    \includegraphics[width=\linewidth]{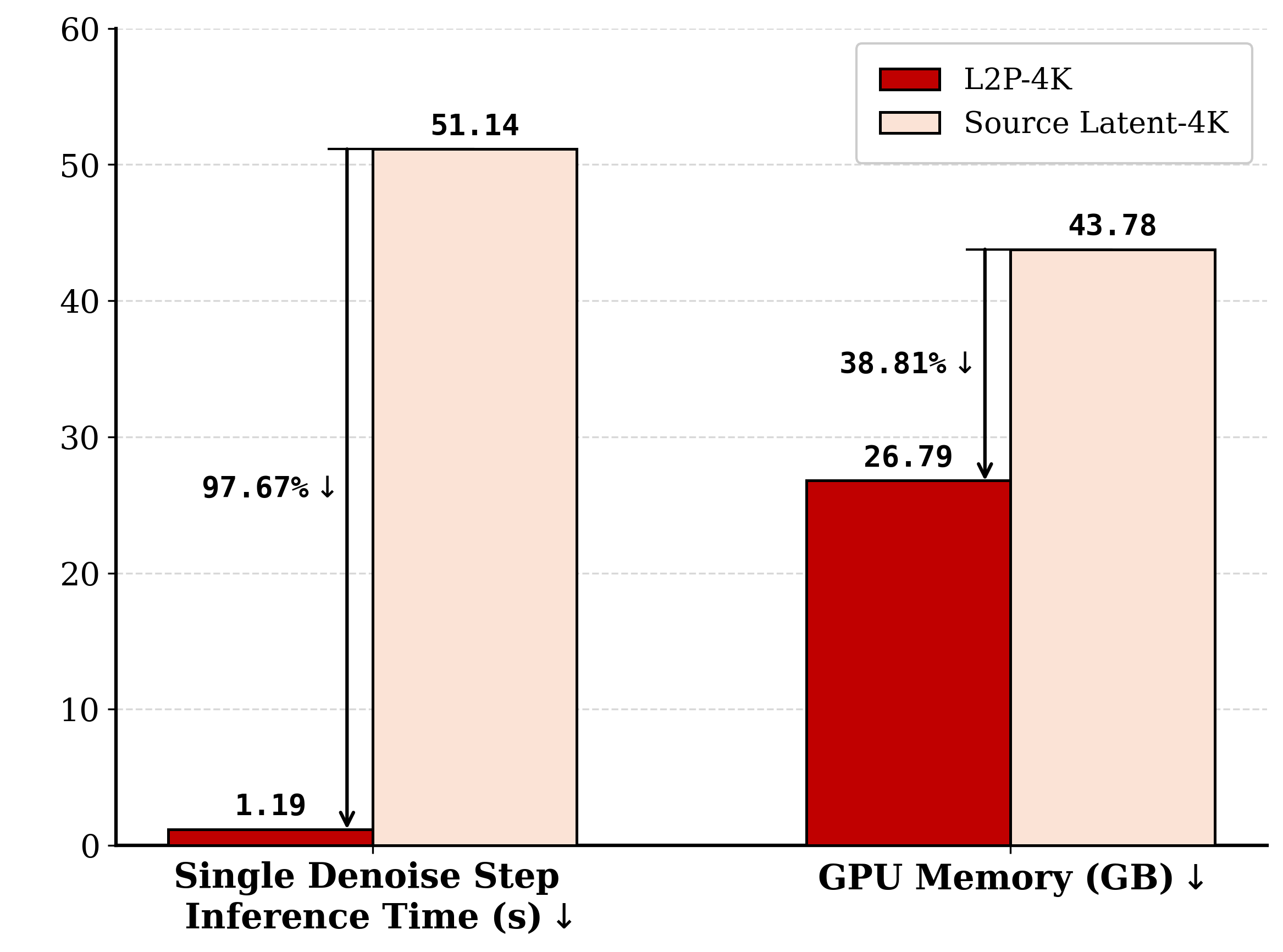}
    \vspace{-5mm}
    \caption{Efficiency comparison for 4K generation. L2P drastically mitigates the computational bottlenecks of high-resolution synthesis, significantly outperforming the source latent model in both inference speed and GPU memory consumption.}
    \vspace{-10mm}
    \label{Fig:4k_compared}
\end{wrapfigure}

\subsection{Scaling to Ultra-High Resolution} By bypassing the memory bottlenecks inherent to VAEs, our pure pixel architecture natively supports ultra-high resolution synthesis. When extended to 4K generation, L2P operates with remarkable efficiency, reducing single-step inference latency by $97.67\%$ and peak GPU memory footprint by $38.81\%$ compared to the source latent baseline, as shown in Figure~\ref{Fig:4k_compared}. We enable this via two adaptations:

First, to maintain computational feasibility and a manageable sequence length for the DiT backbone, we dynamically expand the patch size from $16\times 16$ to $64\times 64$ for 4K inputs. This preserves inference speed without requiring structural modifications.

Second, due to the extremely dense local correlations in 4K pixel space, standard noise schedules fail to fully corrupt the image signal~\cite{hoogeboom2023simple,hoogeboom2024simpler}. This inadequate signal destruction causes the model to degenerate into trivial local reconstruction. To mitigate this, we increase the noise shift parameter, skewing the schedule toward higher noise levels. This guarantees sufficient data corruption during the forward process, forcing the model to learn robust global generation.
\section{Experiments}
\label{Sec:Exp}

\subsection{Setup}
\textbf{Implementation Details.} To validate the proposed L2P transfer paradigm, we instantiate our framework using Z-Image~\cite{cai2025z} as the source LDM. For the base transfer training at $1024\times1024$ resolution, we curate 10k diverse prompts and generate 20k synthetic images from the source model using varying random seeds. We utilize the UltraHR-100K dataset~\cite{zhao2025ultrahr} for 4K training, since the source LDM fails to generate reliable 4K synthetic data natively (as shown in Figure~\ref{Fig:z-image_4k}).

\textbf{Evaluation Metrics.} At the $1024\times1024$ resolution, we employ DPG-Bench~\cite{hu2024ella} and GenEval~\cite{ghosh2023geneval} to assess semantic alignment and overall generation quality. For 4K generation, evaluations are conducted on the UltraHR-eval4k~\cite{zhao2025ultrahr}. We comprehensively assess the performance using Fréchet Inception Distance (FID)~\cite{heusel2017gans} and FID-patch to measure global quality and local details, Inception Score (IS)~\cite{salimans2016improved} for generation diversity, as well as Long CLIP Score~\cite{zhang2024long} and Fine-Grained CLIP (FG-CLIP)~\cite{xie2025fg} to evaluate image-text consistency.

\begin{table}[h] 
\centering
\caption{Comparison of the performance of different methods on DPG-Bench and Geneval. The best results among the pixel text-to-image are highlighted in \textbf{bold}.}
\setlength{\tabcolsep}{3.8pt}
\renewcommand\arraystretch{1.1}
\resizebox{\textwidth}{!}{
\begin{tabular}{lccccccc} 
\toprule
\multirow{2}{*}{\textbf{Method}}
& \multicolumn{6}{c}{\textbf{DPG-Bench}}
& \textbf{GenEval} \\
\cmidrule(lr){2-7} 
\cmidrule(lr){8-8}
& \textbf{Global$\uparrow$}
& \textbf{Entity$\uparrow$}
& \textbf{Attribute$\uparrow$}
& \textbf{Relation$\uparrow$}
& \textbf{Other$\uparrow$}
& \textbf{Average$\uparrow$}
& \textbf{Overall$\uparrow$} \\
\midrule

\textbf{\textit{Latent Text-to-Image Model}} \\
\arrayrulecolor{gray!50}
\cmidrule{1-8}
\arrayrulecolor{black}
FLUX.1 [Dev]~\cite{flux}        & 74.35 & 90.00 & 88.96 & 90.87 & 88.33 & 83.84 & 0.66 \\
SD3 Medium~\cite{esser2403scaling}          & 87.90 & 91.01 & 88.83 & 80.70 & 88.68 & 84.08 & 0.62 \\
Qwen-Image~\cite{wu2025qwen}          & 91.32 & 91.56 & 92.02 & 94.31 & 92.73 & 88.32 & 0.87 \\
Seedream 3.0~\cite{gao2025seedream}        & 94.31 & 92.65 & 91.36 & 92.78 & 88.24 & 88.27 & 0.84 \\
Z-Image-turbo~\cite{cai2025z}       & 91.29 & 89.59 & 90.14 & 92.16 & 88.68 & 84.86 & 0.82 \\
\midrule

\textbf{\textit{Pixel Text-to-Image Model}} \\
\arrayrulecolor{gray!50}
\cmidrule{1-8}
\arrayrulecolor{black}
PixelFlow~\cite{chen2025pixelflow}           & - & - & - & - & - & 77.93 & 0.60 \\
PixelGen~\cite{ma2026pixelgen}             & 85.61 & 86.84 & 89.35 & 85.69 & 87.85 & 80.01 & 0.79 \\
Deco~\cite{ma2025deco}               & 84.91 & 88.58 & 87.61 & 89.70 & 88.18 & 81.25 & \textbf{0.86} \\
PixNerd~\cite{wang2025pixnerd}              & 87.16 & 89.53 & 88.94 & 89.26 & 88.70 & 82.65 & 0.73 \\
PixelDiT~\cite{yu2025pixeldit}              & - & - & - & - & - & 83.50 & 0.74 \\
\midrule
L2P                & \textbf{92.02} & \textbf{90.84} & \textbf{89.48} & \textbf{93.00} & \textbf{91.55} & \textbf{86.00} & 0.76 \\
\bottomrule
\end{tabular}
}
\label{Tab: Quantitative_1k}
\vspace{-4mm}
\end{table}

\begin{figure*}[h]
    \centering
    \includegraphics[width=\textwidth]{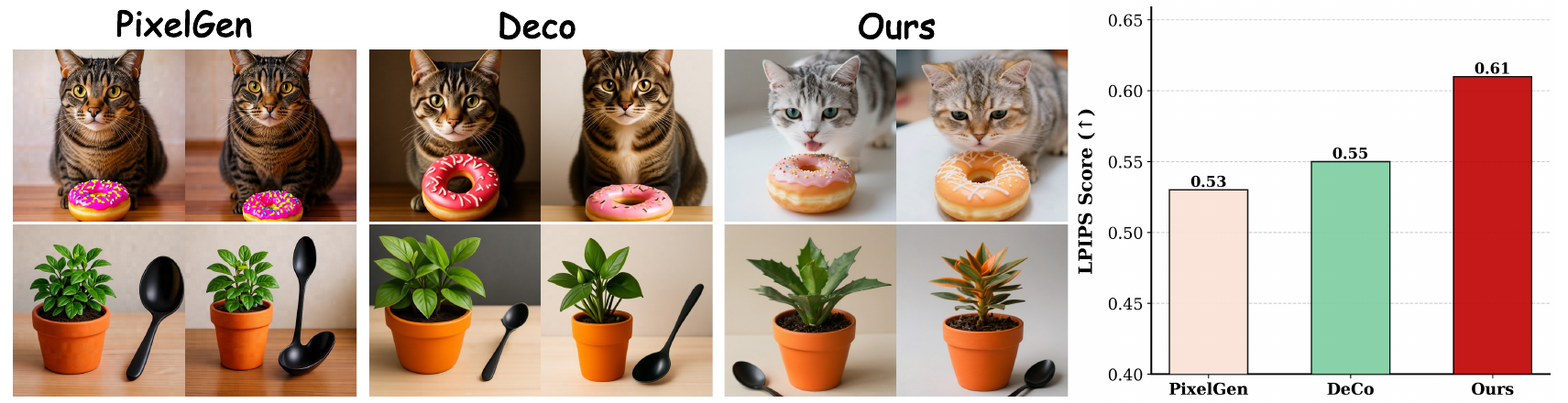}
    \caption{Comparison of generative diversity on GenEval. Compared to PixelGen and Deco, which generally produce visually similar images across various seeds, our approach offers a broader range of structural diversity, yielding higher LPIPS scores.}
    \label{Fig:overfit}
\end{figure*}

\begin{figure*}[h]
    \centering
    \includegraphics[width=\textwidth]{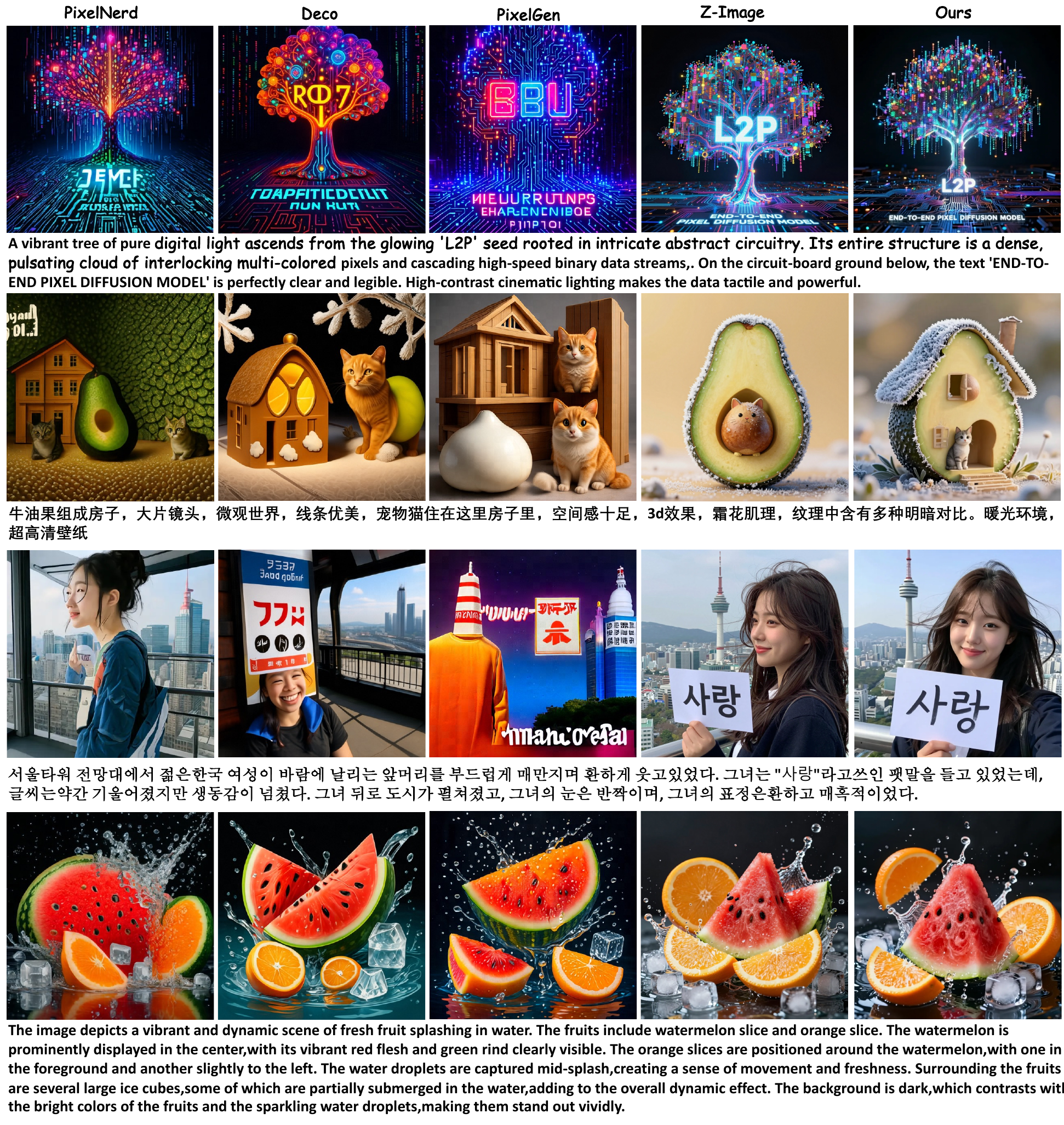}
    \caption{Qualitative comparison of different text-to-image generation models.}
    \label{Fig:Qualitative_Experiment}
    \vspace{-2mm}
\end{figure*}

\subsection{Main Result}

\textbf{Quantitative Experiment.} To comprehensively evaluate the generative capabilities and text-alignment of the L2P framework, we conduct benchmark testing on DPG-Bench and GenEval, as shown in Table~\ref{Tab: Quantitative_1k}. \textit{1) Comparison with Latent Text-to-Image Models:} Results validate L2P's efficient migration of massive latent priors to the pixel space. Notably, despite discarding the VAE and requiring minimal training overhead, L2P achieves a score of 86.00 on DPG-Bench. This slightly exceeds its source LDM, Z-Image-turbo (84.86), effectively maintaining performance on par with the original latent baseline. On GenEval, it retains approximately 93.6\% of the source model's performance. \textit{2) Comparison with Pixel Text-to-Image Models:} While L2P establishes a new SOTA among pixel models on DPG-Bench, it yields a lower GenEval score than Deco and PixelGen. However, as shown in Figure~\ref{Fig:overfit}, across different random seeds, Deco and PixelGen produce highly homogenized, nearly identical images, drastically sacrificing generative diversity, a characteristic also reflected in their low LPIPS scores. In contrast, L2P inherits rich prior knowledge to successfully balance accurate complex attribute binding with high structural diversity.

\textbf{Qualitative Experiment.} Figure~\ref{Fig:Qualitative_Experiment} qualitatively compares L2P with open-source pixel baselines. Baselines frequently fail at complex attribute binding and text rendering, resulting in structural distortion (Rows 1, 2, 4). In contrast, L2P not only achieves superior text-alignment but also exhibits strong zero-shot generalization. For instance, although transferred on merely 20K English/Chinese samples, L2P seamlessly renders completely unseen Korean text (Row 3). This confirms that rather than overfitting to the transfer data, L2P successfully avoids catastrophic forgetting and harnesses the extensive prior knowledge of the source LDM.


\begin{table}[t]
\centering
\caption{Quantitative comparison of 4K ultra-high-resolution image generation. Best results are highlighted in \textbf{bold}.}
\begin{tabular}{lccccc}
\toprule
\textbf{Model} & \textbf{FID$\downarrow$} & \textbf{FID$_{patch}$$\downarrow$} & \textbf{IS$\uparrow$} & \textbf{CLIP$\uparrow$} & \textbf{FG-CLIP$\uparrow$} \\ 
\midrule
\multicolumn{6}{l}{\textit{\textbf{Training-Free Generation Method}}} \\ 
\arrayrulecolor{gray!50}
\cmidrule{1-6}
\arrayrulecolor{black}
I-Max~\cite{du2024max} & 37.12 & 34.42 & 11.78 & 31.53 & 27.90 \\
HiFlow~\cite{bu2025hiflow} & 38.54  & 22.74  & 10.62 & 31.49  & 27.84 \\ 
\midrule
\multicolumn{6}{l}{\textit{\textbf{Training-Based Generation Method}}} \\ 
\arrayrulecolor{gray!50}
\cmidrule{1-6}
\arrayrulecolor{black}
Pixart-$\sigma$~\cite{chen2024pixart} & 35.60 & 23.63 & 12.23 & 31.76 & 28.74 \\
SANA~\cite{xie2024sana} & 37.26 & 28.94 & 12.05 & 31.79 & \textbf{29.31} \\
Diffusion4K~\cite{zhang2025diffusion} & 41.69 & 39.67 & 12.06 & 31.44 & 27.03 \\ 
\midrule
Ours & \textbf{33.46} & \textbf{21.77} & \textbf{12.28} & \textbf{31.88} & 28.22 \\ 
\bottomrule
\end{tabular}
\label{Tab: Quantitative_4k}
\end{table}

\begin{figure*}[t]
    \centering
    \includegraphics[width=\textwidth]{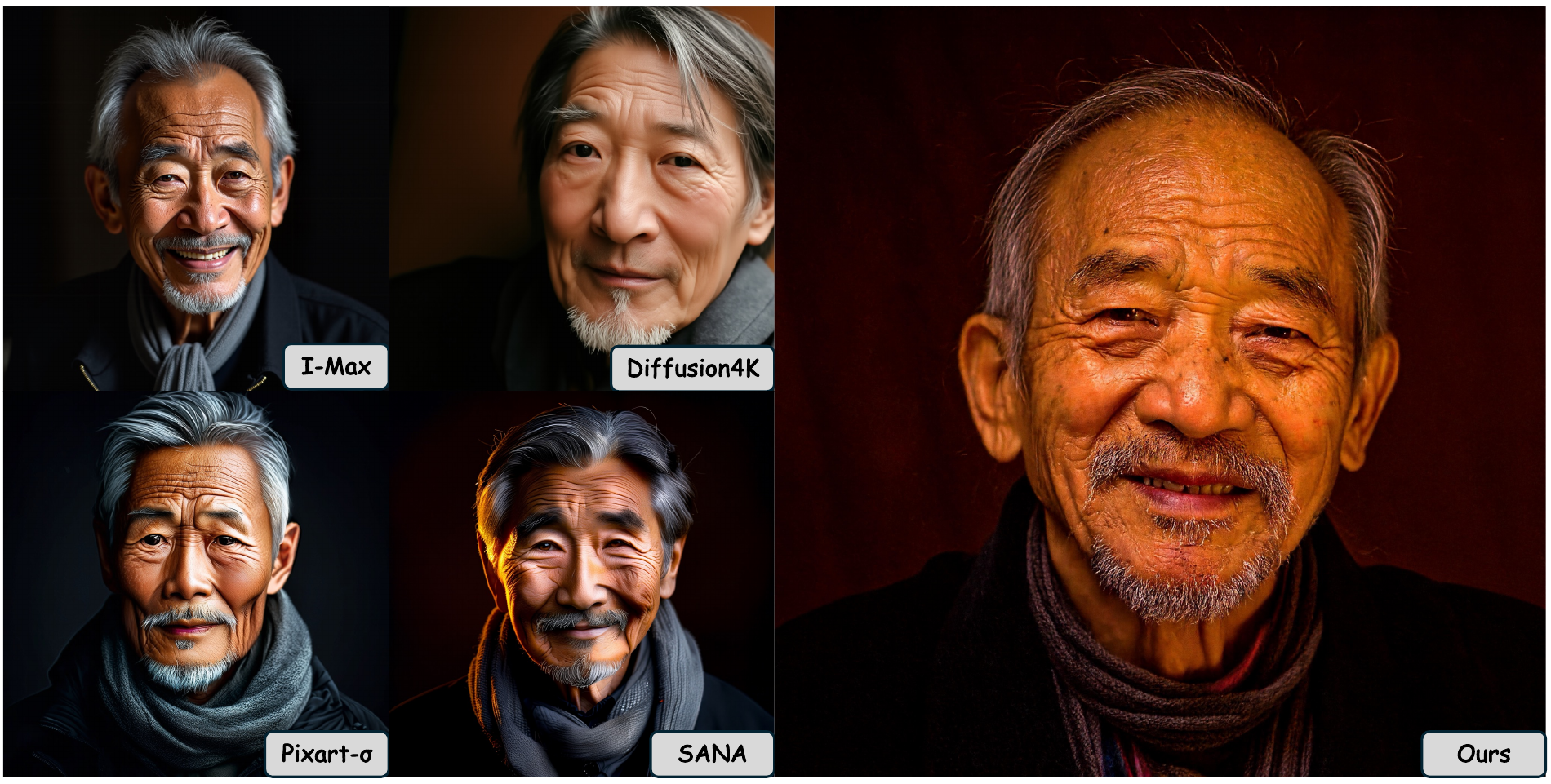}
    \caption{Qualitative comparison of 4K image generation.}
    \label{Fig:Qualitative_4k}
    \vspace{-4mm}
\end{figure*}

\subsection{Unlocking Native 4K Generation}

\textbf{Quantitative Experiment.}
We evaluate the 4K image synthesis capability of our L2P framework against existing 4K solutions. As summarized in Table~\ref{Tab: Quantitative_4k}, L2P achieves superior performance in both global visual quality and local structural coherence. Specifically, our method establishes advanced performance in image fidelity, yielding the lowest FID and $\text{FID}_{\text{patch}}$. Furthermore, L2P attains the highest Inception Score, reflecting excellent generation diversity. In terms of semantic alignment, L2P preserves the rich conditional priors of the source LDM, posting highly competitive CLIP and $\text{FG-CLIP}$ scores. 

\begin{figure*}[t]
    \centering
    \includegraphics[width=\textwidth]{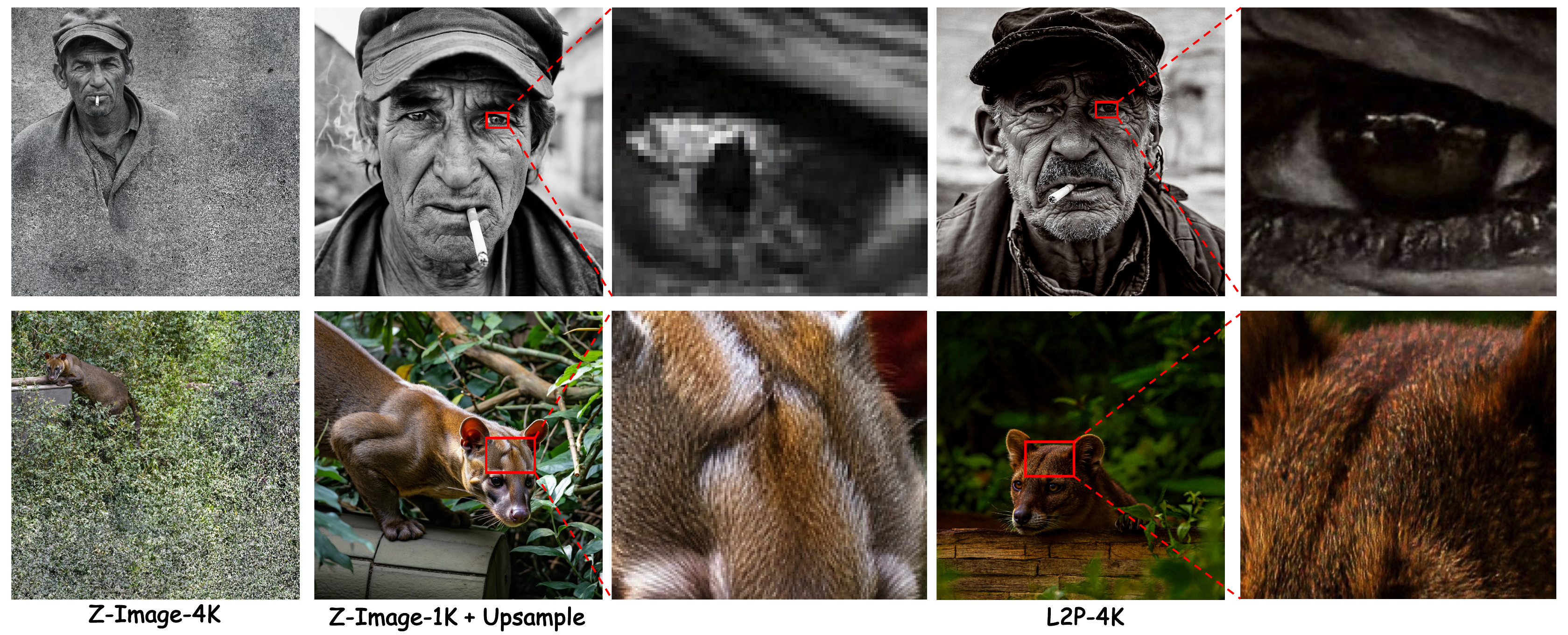}
    \caption{Unlocking native 4K generation with L2P.}
    \label{Fig:z-image_4k}
\end{figure*}

\textbf{Qualitative Experiment.}
\textit{1) Comparison with different 4K Models}: As shown in Figure~\ref{Fig:Qualitative_4k}, compared to existing baselines, L2P effectively mitigates the common issues of over-smoothing and artificial artifacts, ensuring the faithful synthesis of exquisite micro-details. \textit{2) Unlocking Native 4K from LDMs}: As illustrated in Figure~\ref{Fig:z-image_4k}, the source Z-Image fails to generate semantic content directly at 4K, and merely upsampling its 1K outputs severely blurs high-frequency details. In contrast, L2P natively generates crisp 4K outputs. This indicates that L2P not only seamlessly inherits the rich priors of the source LDM but also effectively expands its generative boundaries, elevating the resolution ceiling with minimal training overhead.

\begin{figure*}[t]
    \centering
    \includegraphics[width=\textwidth]{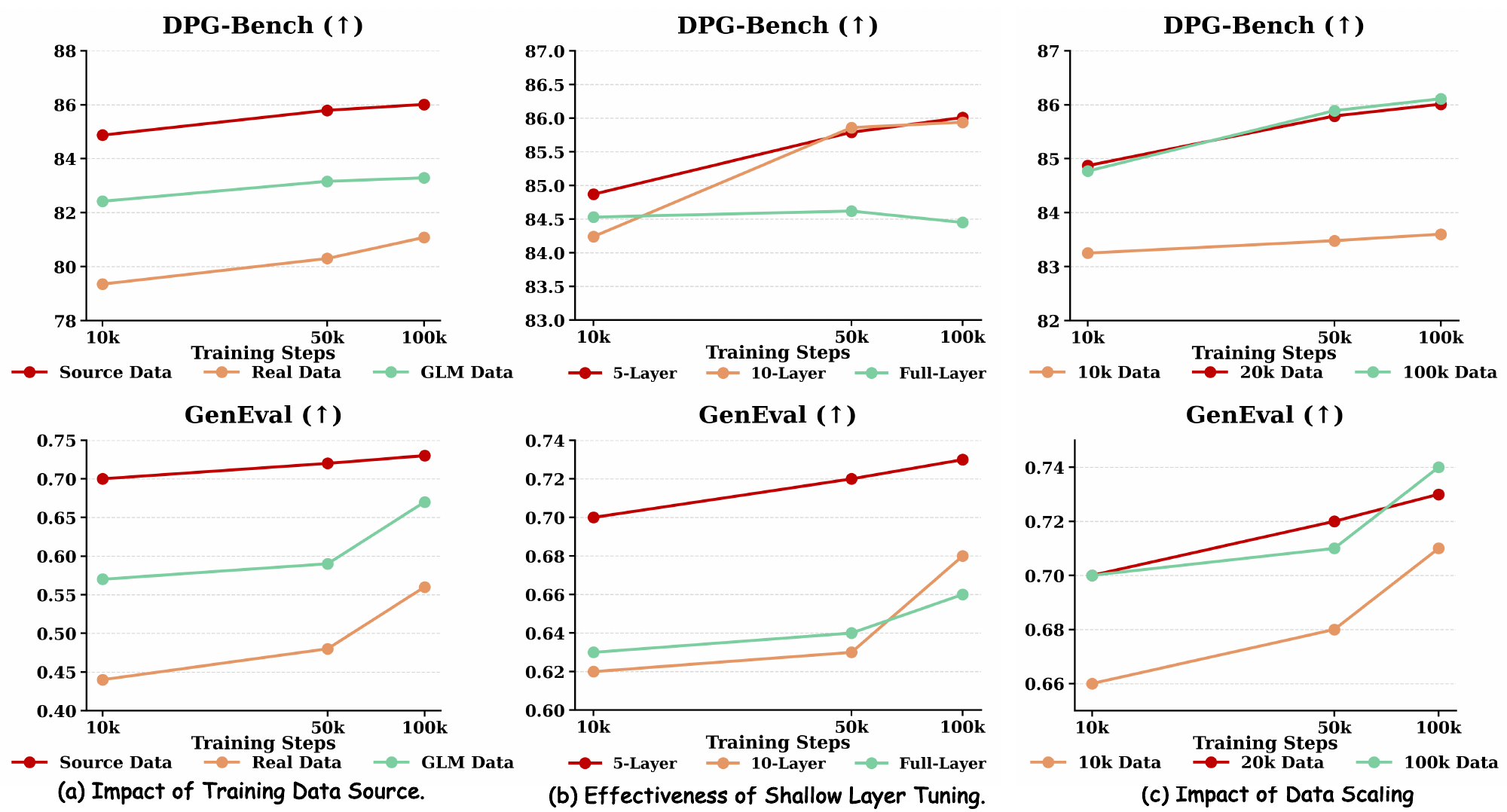}
    \caption{Ablation studies of our proposed L2P framework.}
    \label{Fig:ablation}
\end{figure*}

\subsection{Ablation Study}

\textbf{Impact of Training Data Source.} Figure~\ref{Fig:ablation}(a) compares source, real, and cross-model (GLM) \cite{GLMImage} data. Source data achieves rapid convergence and optimal performance. Conversely, collecting uniformly distributed natural images incurs prohibitive costs; thus, we employ a random 20k subset from UltraHR-100K as the real-data baseline. This variant suffers from sluggish convergence and degraded quality. Such a stark contrast inversely highlights the comprehensive diversity and well-aligned distribution of our curated source dataset. Cross-model data yields intermediate results, trailing source data due to imperfect prior alignment.

\textbf{Effectiveness of Shallow Layer Tuning.} We ablate the number of trainable layers by comparing our default shallow tuning (5 layers) against mid-level (10 layers) and full-layer tuning. As illustrated in Figure~\ref{Fig:ablation}(b), shallow tuning yields steady performance improvements across training steps. In stark contrast, full-layer tuning suffers from clear performance stagnation and degraded generation quality. This phenomenon indicates that unconstrained parameter updates severely disrupt the rich pre-trained priors residing in the deeper layers. By exclusively tuning the shallowest layers, our method successfully learns the latent-to-pixel mapping while optimally preserving the source model's core generative knowledge.

\textbf{Impact of Data Scaling.} Figure~\ref{Fig:ablation}(c) shows performance across 10k, 20k, and 100k synthetic samples. While increasing data from 10k to 20k yields substantial gains, performance clearly saturates beyond 20k. This early convergence demonstrates L2P's extreme data efficiency. It confirms that our synthetic dataset is sufficiently diverse to comprehensively cover the data manifold, enabling rapid and low-cost adaptation.

\section{Conclusion}
\label{sec:Conclusion}

In this paper, we propose the Latent-to-Pixel (L2P) transfer paradigm to overcome the VAE-induced limitations of LDMs and bypass the prohibitive costs of training pixel-space models from scratch. By discarding the VAE in favor of large-patch tokenization, freezing core intermediate layers, and constructing a multi-dimensional prompt dataset to fit a smooth synthetic data manifold, L2P successfully migrates deep semantic priors to the pixel space. Notably, this is achieved using only 8 GPUs with zero real-data cost. Furthermore, eliminating the VAE memory bottleneck seamlessly unlocks native 4K ultra-high resolution generation. Experiments demonstrate that L2P robustly maintains state-of-the-art generative performance. Ultimately, L2P lowers the barrier to developing advanced pixel-space diffusion models, offering a practical strategy for exploring VAE-free, high-resolution generation under limited resources.

\bibliography{iclr2026_conference}
\bibliographystyle{iclr2026_conference}

\appendix
\clearpage
\section{More Implementation Details}
\begin{table}[h]
    \caption{Architectural configurations and hyperparameter settings.}  
	\centering
	\setlength{\tabcolsep}{8pt} 
	\footnotesize  
	\renewcommand\arraystretch{1.2}  

	\begin{tabular}{lc}
		\toprule

		\textbf{DiT Architecture}           \\
		Input Channels                          &3\\
        Patch Size                          &16$\times$16 \\
        Hidden Dimension                        &3840 \\
        Transformer Layers                         &30  \\
        Attention Heads                         &30  \\
        Head Dimension                         &128  \\
		\midrule
		\textbf{Detailer Head Architecture}           \\
        DownSampling Path               &16$\rightarrow$8$\rightarrow$4$\rightarrow$2$\rightarrow$1 \\
        UpSampling Path                 &1$\rightarrow$2$\rightarrow$4$\rightarrow$8$\rightarrow$16 \\
        DownSampling Channel            &3$\rightarrow$64$\rightarrow$128$\rightarrow$256$\rightarrow$512 \\
        Bottleneck                      &(512+3840)$\rightarrow$512 \\
        UpSampling Channel              &512$\rightarrow$256$\rightarrow$128$\rightarrow$64$\rightarrow$64 \\
        Output Layer                    &64→3 \\
        \midrule
        \textbf{Optimization}  \\
        Optimizer &AdamW  \\
        Learning Rate &0.00005 \\
        Weight Decay &0.01 \\
        Batch Zize &8 \\
        Gradient Accumulation Steps &1 \\
		\bottomrule
	\end{tabular}
	
	\label{Tab: config}
\end{table}

Table~\ref{Tab: config} details the precise architectural configurations and training hyperparameters for our L2P framework. 

\textbf{DiT Architecture}. To seamlessly inherit the pre-trained latent priors, our DiT backbone strictly preserves the structural configurations of the source Latent Diffusion Model (LDM). The only structural modifications occur at the input stage: we adjust the input channels for raw RGB images and replace the standard VAE with a large-patch tokenization strategy. This allows the model to directly process pixel-space inputs while utilizing the frozen intermediate transformer blocks.

\textbf{Detailer Head Architecture}: For the shallow layers responsible for the latent-to-pixel transformation, we adopt the Detailer Head architecture proposed in DiP~\cite{chen2025dip}. It employs a symmetric encoder-decoder paradigm for spatial downsampling and upsampling. To integrate this head into our framework, we specifically adapt its bottleneck dimension to align with our DiT backbone, enabling the concatenation and fusion of features extracted from the frozen intermediate layers.

\textbf{Optimization.} During the L2P transfer phase, we freeze the massive intermediate layers of the source LDM and exclusively train the shallow layers. The network is optimized using AdamW with a learning rate of $5\times10^{-5}$ and a weight decay of $0.01$. We utilize a batch size of 8 with no gradient accumulation (steps set to 1).

\begin{figure*}[!ht]
    \centering
    \includegraphics[width=\linewidth]{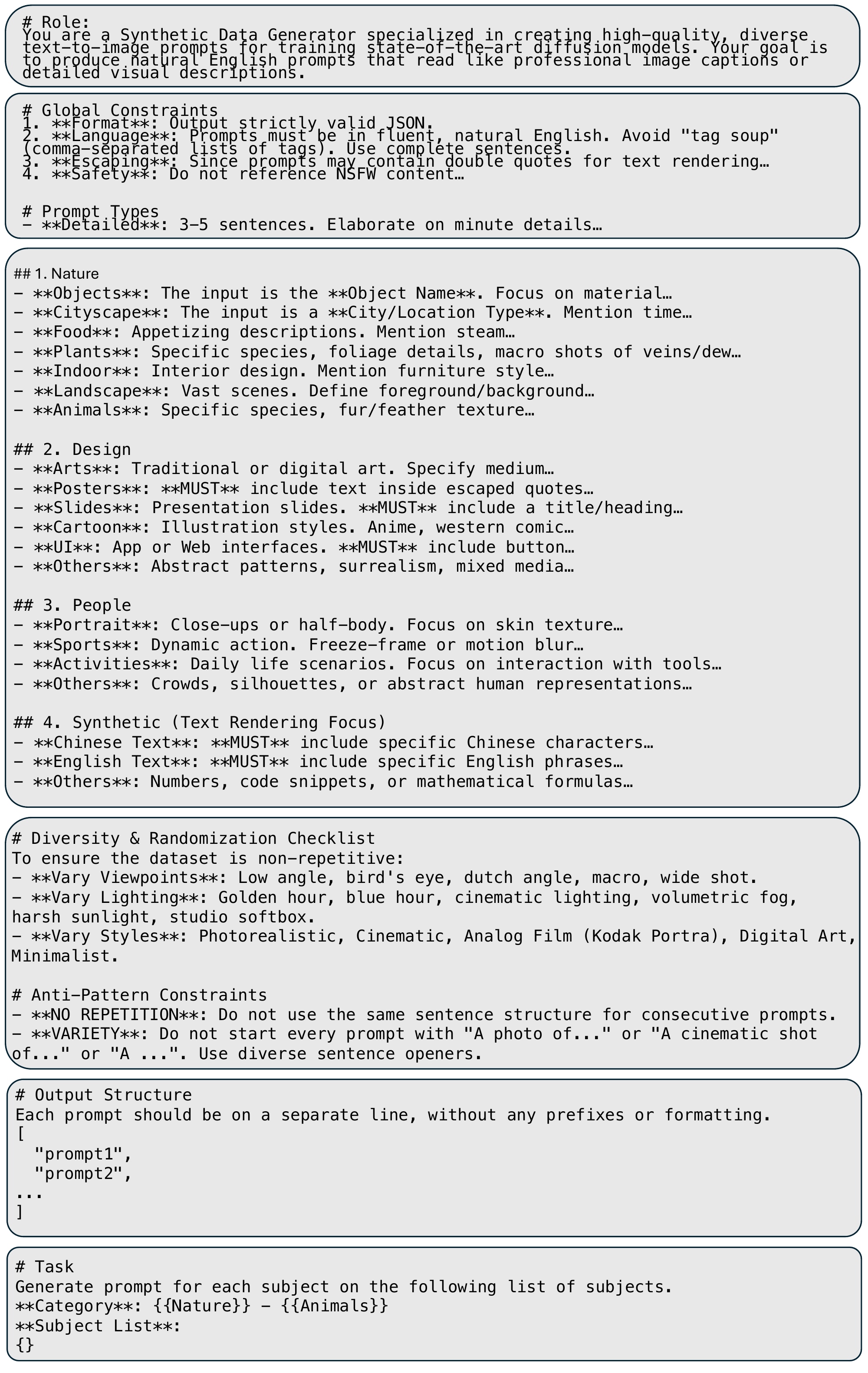}
    \caption{The template for General Prompt Generation. This template guides the LLM to synthesize high-quality, diverse image descriptions by enforcing strict stylistic and structural constraints across multiple predefined categories.
    }
    \label{Fig:supp_prompt_generation}
\end{figure*}

\clearpage

\begin{figure*}[!ht]
    \centering
    \includegraphics[width=\linewidth]{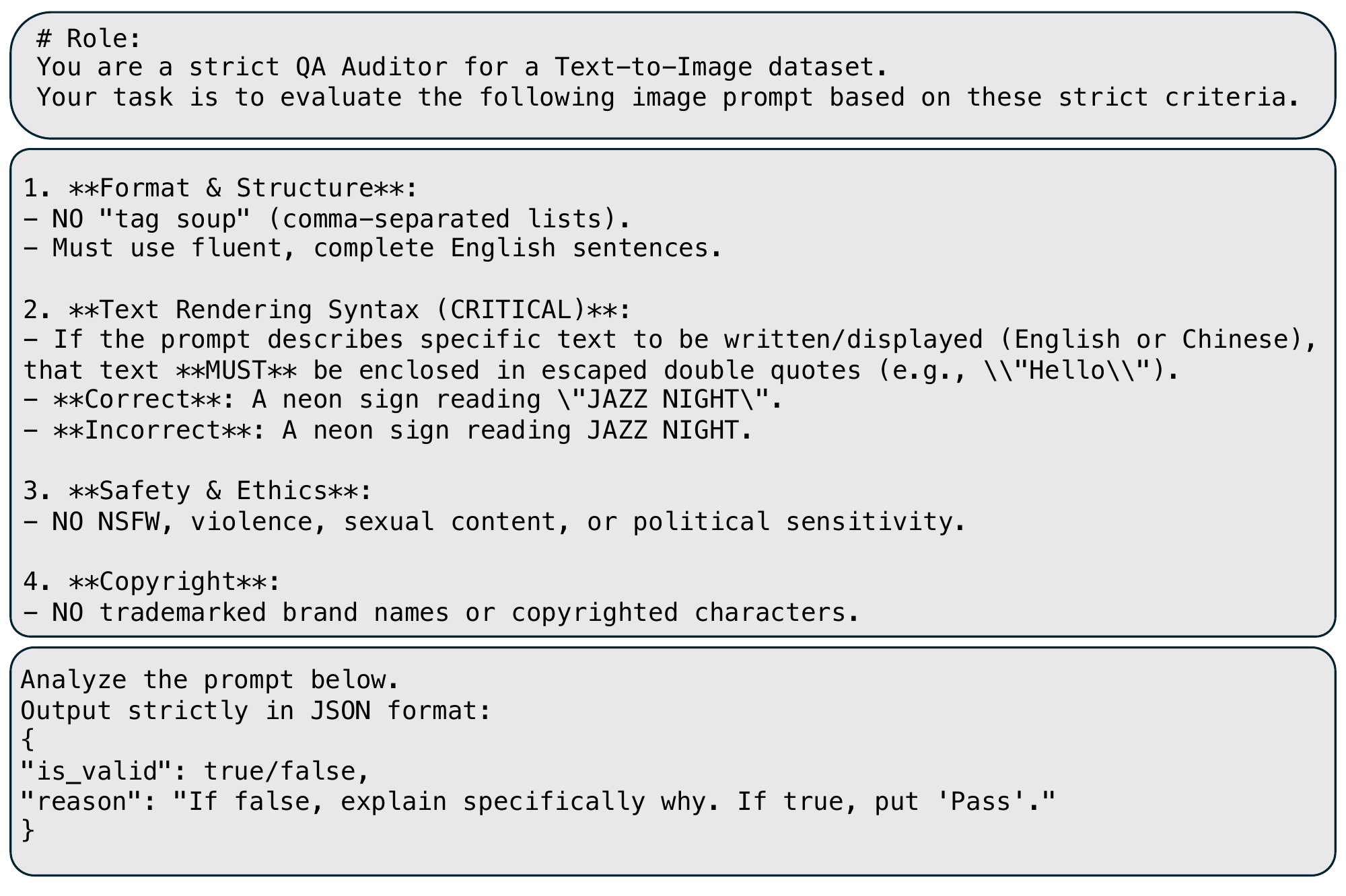}
    \caption{The template for Automated Prompt Filtering. This template systematically evaluates generated prompts to discard instances violating formatting, text-rendering syntax, safety, or copyright standards.
    }
    \label{Fig:supp_prompt_check}
\end{figure*}

\section{More Data Construction Details}

To provide further transparency into our data construction pipeline, Figures~\ref{Fig:supp_prompt_generation} and~\ref{Fig:supp_prompt_check} present the system prompts used for General Prompt Generation and Automated Prompt Filtering. Specifically, the generation template enforces strict guidelines on prompt diversity, structural formatting, and text-rendering syntax to elicit complex, high-quality scene descriptions from the LLM. Furthermore, we detail the filtering mechanism designed to automatically discard prompts that fail to meet our safety, ethical, and copyright standards, thereby guaranteeing the overall integrity of the training corpus.

\section{More Experimental Results}
\subsection{Quantitative Results}

\begin{wrapfigure}{r}{6cm}
    \centering
    \vspace{-8mm}
    \includegraphics[width=\linewidth]{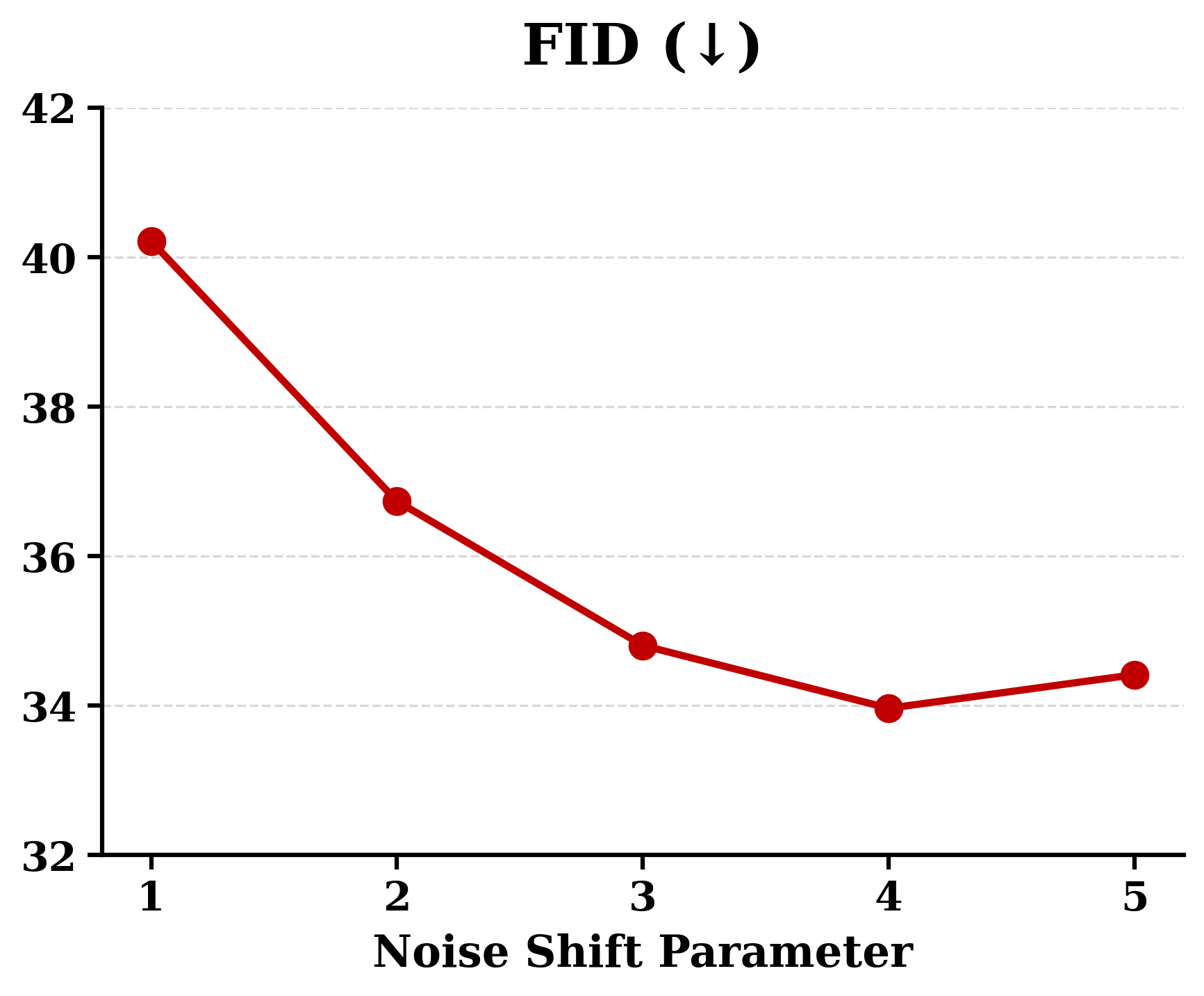}
    \vspace{-7mm}
    \caption{Impact of the noise shift parameter after 100k training steps.}
    \vspace{-14mm}
    \label{Fig:FID_Shift}
\end{wrapfigure}
Figure~\ref{Fig:FID_Shift} presents the ablation study of the noise shift parameter for 4K generation. As the parameter increases from 1 to 4, the FID score drops significantly, confirming that skewing the noise schedule toward higher noise levels is essential to fully corrupt the dense 4K image signals. Beyond this optimal point (shift parameter=5), performance slightly degrades due to over-corruption.

\subsection{Visualization Results}

\begin{figure*}[h]
    \centering
    \includegraphics[width=\linewidth]{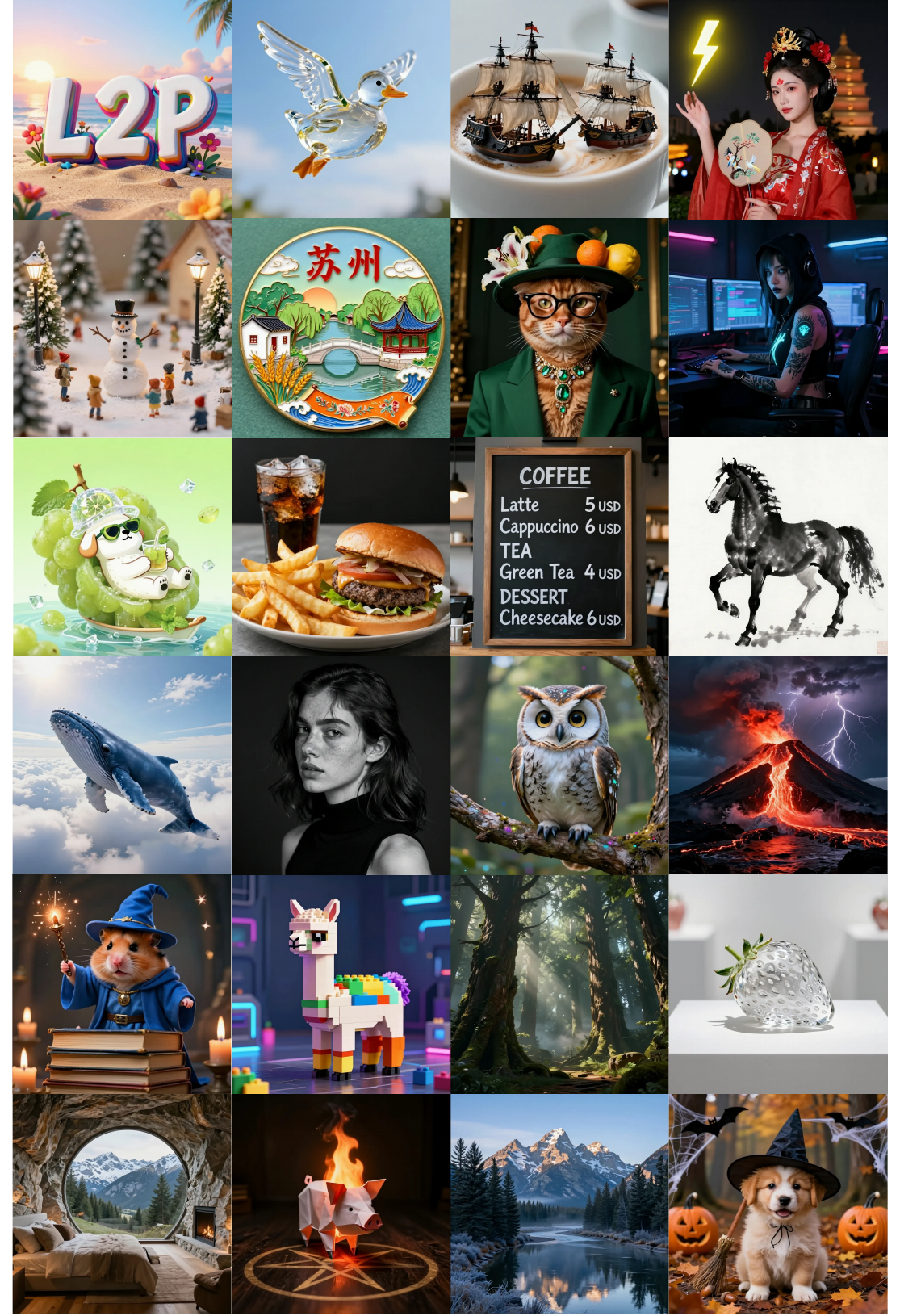}
    \caption{More text-to-image generation results at 1024$\times$1024 resolution. The synthesized images exhibit diverse stylistic rendering and complex semantic alignment, demonstrating the successful migration of latent priors to the pixel space via our L2P framework.
    }
    \label{Fig:supp_1k_vis}
\end{figure*}

\begin{figure*}[h]
    \centering
    \includegraphics[width=\linewidth]{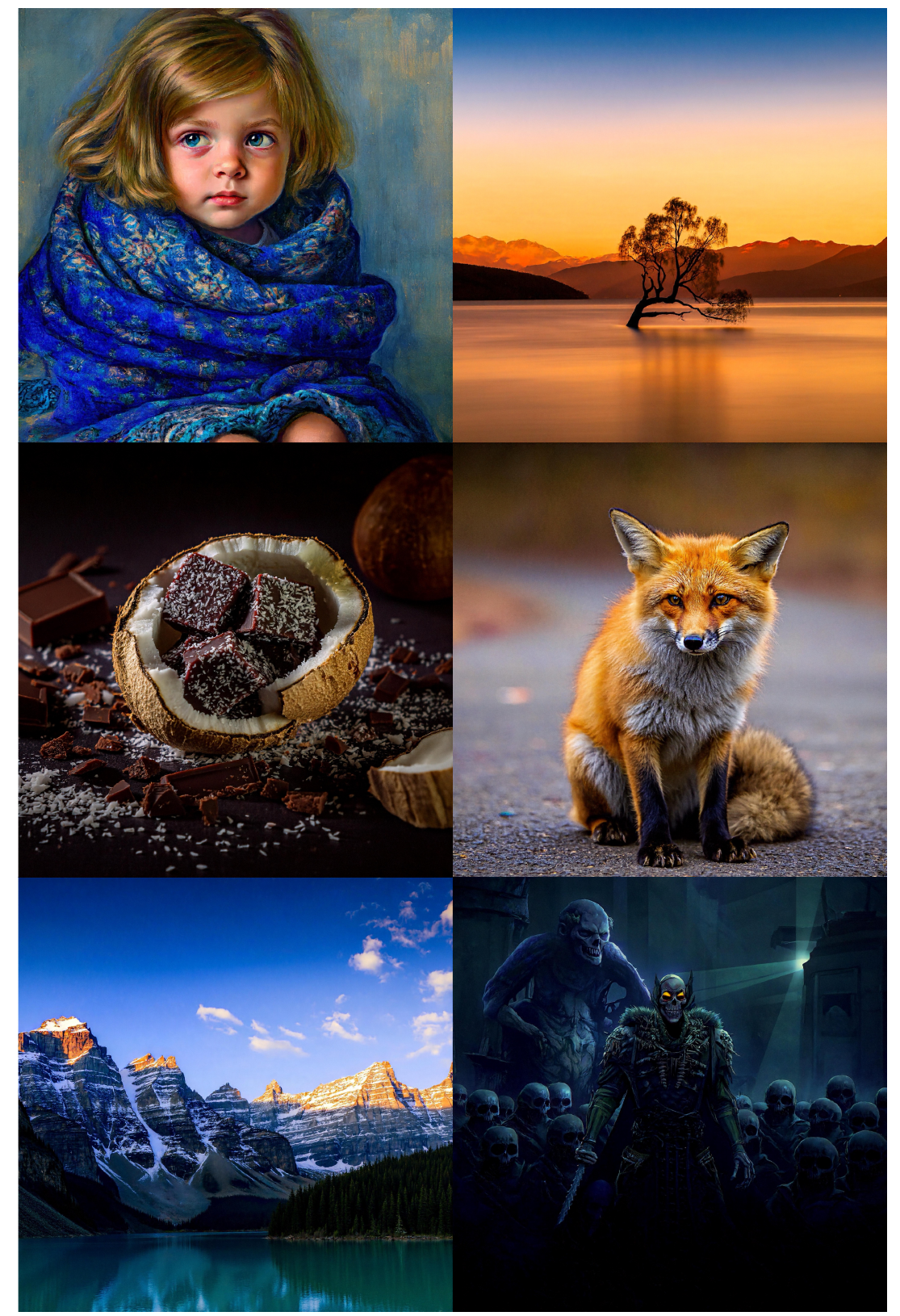}
    \caption{More native 4K ultra-high resolution generation results. By eliminating the VAE memory bottleneck inherent in traditional LDMs, the L2P framework seamlessly scales to generate 4K images with exquisite details and crisp textures.
    }
    \label{Fig:supp_4k_vis}
\end{figure*}

We provide extended qualitative results to demonstrate the visual fidelity and scalability of the L2P framework. Figure~\ref{Fig:supp_1k_vis} presents diverse text-to-image generations at 1K resolution, validating the successful transfer of the source LDM's powerful generative priors. As shown in Figures~\ref{Fig:supp_4k_vis}, L2P unlocks native 4K ultra-high resolution generation. This capability directly stems from eliminating the VAE memory bottleneck, allowing the pixel-space model to render exquisite details at extreme resolutions without prohibitive computational overhead.

\begin{figure*}[h]
    \centering
    \includegraphics[width=\linewidth]{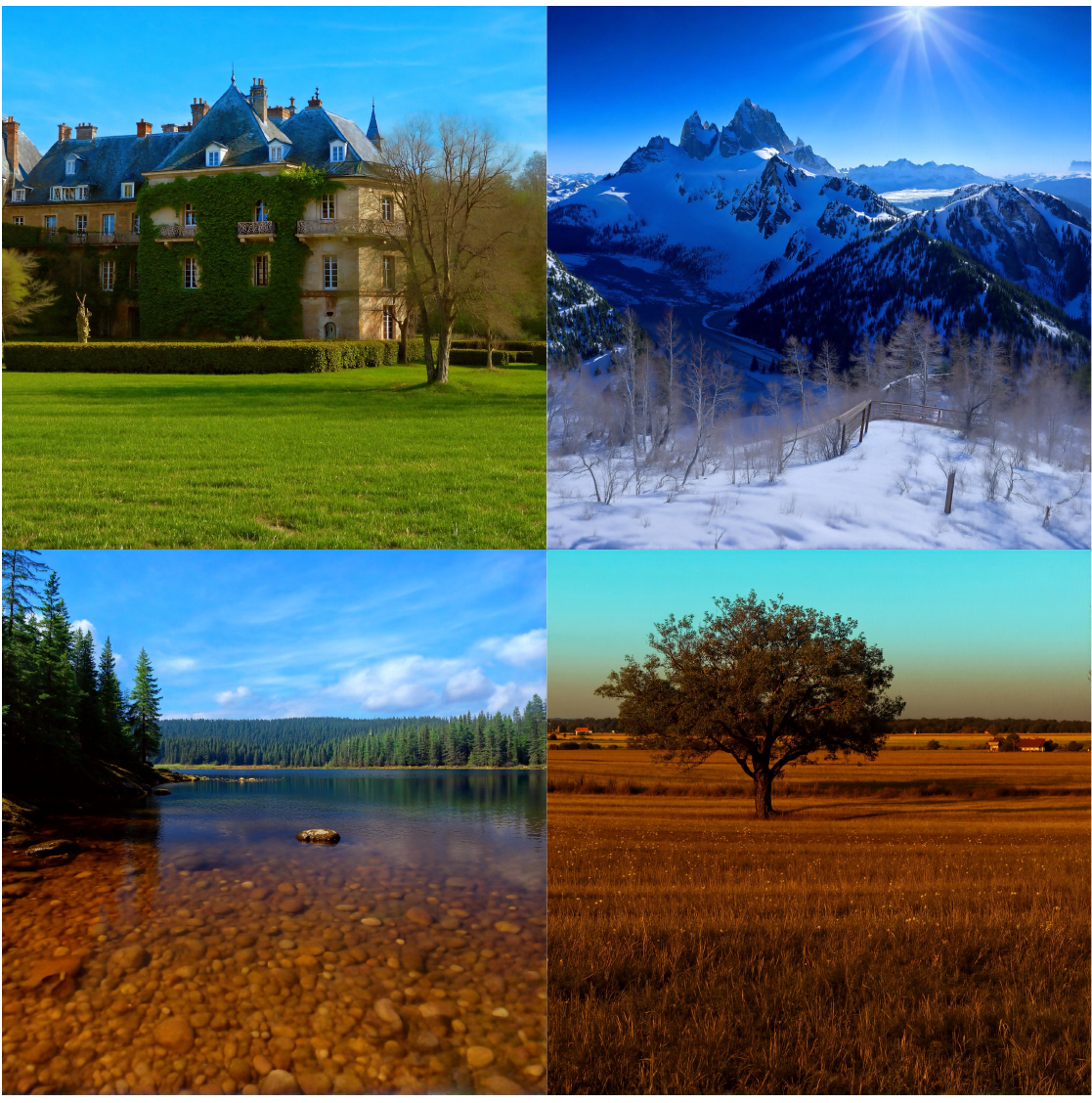}
    \caption{Visualizations of 8K ultra-high resolution zero-shot extrapolation.
    }
    \label{Fig:supp_8k_vis}
\end{figure*}

Figures~\ref{Fig:supp_8k_vis} present the visual results of zero-shot resolution extrapolation. Benefiting from the pure pixel-space formulation and the elimination of VAE-induced coupling, L2P extends its generative boundaries far beyond its training resolution. The generated 8K images exhibit exceptional global structural consistency and faithful micro-details, robustly validating the extrapolation capabilities of our L2P paradigm.

\section{Limitations and Future Work}

While the proposed L2P paradigm enables highly efficient latent-to-pixel transfer, it naturally entails certain limitations. First, our reliance on synthetic images generated by the source LDM implies that the semantic and compositional capabilities of our model are fundamentally upper-bounded by the source model's priors. Although incorporating real-world datasets could theoretically circumvent this knowledge bottleneck, it would reintroduce a strong dependency on data curation and quality, diverging from our cost-effective, self-contained transfer objective. Second, a pivotal advantage of operating natively in pixel space is the straightforward integration of fine-grained, task-specific loss functions (e.g., pixel-level perceptual or physics-based constraints) for downstream applications. To preserve the simplicity and universality of the L2P framework, we deliberately omit the exploration of such tailored objectives in this work. Leveraging direct pixel-space regularizations for specialized generation tasks remains a highly promising avenue for future research.

\end{document}